%% file: main.tex
\documentclass[a4paper,fleqn]{article}
\usepackage{INTERSPEECH2021,amssymb,amsmath,epsfig}
\usepackage{xspace}
\usepackage{srcltx}
\usepackage{caption}
\usepackage{subcaption}
\usepackage[export]{adjustbox}
\usepackage{multirow}
\ninept
\def\reg{{\rm\ooalign{\hfil
     \raise.07ex\hbox{\scriptsize R}\hfil\crcr\mathhexbox20D}}}

\usepackage{color}
\usepackage[dvipsnames]{xcolor}
\usepackage{makecell}
\usepackage{refcount}

\newcommand{\CMT}[1]{{}}

\def\reg{{\rm\ooalign{\hfil
     \raise.07ex\hbox{\scriptsize R}\hfil\crcr\mathhexbox20D}}}

\ninept
\usepackage{makecell}
\begin{document}
\title{Tied \& Reduced RNN-T Decoder}
\name{Rami Botros, Tara N. Sainath, Robert David, Emmanuel Guzman, Wei Li, Yanzhang He}
\address{Google Inc., U.S.A}
\email{\{ramibotros, tsainath, lrdx, emmanuelguzman, mweili, yanzhanghe\}@google.com}
\interfootnotelinepenalty=10000
\maketitle
\input{abstract}
\input{intro}

\input{rnnt}

\input{architecture}

\input{experiments}
\input{results}
\input{conclusions}
\section{Acknowledgements}
The authors thank Trevor Strohman, Ehsan Variani and Shankar Kumar for helpful discussions on embedding decoders.
\newpage
\bibliographystyle{IEEEbib}
\bibliography{main}
\end{document}

%% file: abstract.tex
\begin{abstract}
Previous works on the Recurrent Neural Network-Transducer (RNN-T) models have shown that, under some conditions, it is possible to simplify its prediction network with little or no loss in recognition accuracy \cite{Variani20, ghodsi2020rnn, prabhavalkar2020less}. This is done by limiting the context size of previous labels and/or using a simpler architecture for its layers instead of LSTMs.  The benefits of such changes include reduction in model size, faster inference and power savings, which are all useful for on-device applications.

In this work, we study ways to make the RNN-T decoder (prediction network + joint network) smaller and faster without degradation in recognition performance. Our prediction network performs a simple weighted averaging of the input embeddings, and shares its embedding matrix weights with the joint network's output layer (a.k.a. weight tying, commonly used in language modeling \cite{inan2016tying}). This simple design, when used in conjunction with additional Edit-based Minimum Bayes Risk (EMBR) training, reduces the RNN-T Decoder from 23M parameters to just 2M, without affecting word-error rate (WER). 
\end{abstract}
\noindent\textbf{Index Terms}: end-to-end, speech recognition, on-device, limited memory

%% file: intro.tex
\section{Introduction \label{sec:introduction}}

Research on end-to-end (E2E) models has produced promising results on various speech recognition tasks ~\cite{graves2012sequence, chan2015listen, bahdanau2016end, chiu2018state, he2019streaming, li2020comparison, sainath2020streaming}. These models attempt to simultaneously learn the acoustic, pronunciation and language models of conventional speech recognition systems using a single neural network. In addition to being simpler to train, they are usually much smaller in size compared to the conventional systems~\cite{pundak2016lower}, making them suitable for on-device applications~\cite{he2019streaming, li2019improving, sainath2020streaming}. On-device E2E models have the potential to improve privacy and reduce recognition latency, especially if a streaming model, such as RNN-T, is used.

There are research efforts to bring RNN-T to edge devices. Some relevant challenges are a requirement for low latency, and dealing with memory constraints. In particular, if the complete model cannot completely fit in-memory on hardware accelerators, smaller partitions of it need to be loaded and processed sequentially, with each such step having a high fixed cost. Hence, smaller models can potentially lead to immense speedup during inference. To this end, we focus on studying methods that make the RNN-T decoder, which comprises the prediction network and joint network, as small and computationally cheap as possible without sacrificing WER performance.

There have been studies in the literature on reducing the complexity of the prediction network, or of the length of input that it is given. In \cite{Variani20}, the input to the prediction network's LSTM is restricted to two history phonemes without observing WER degradation. It is hence suggested that, after training, the LSTM be converted to a fast lookup table of size $|V|^2$, where $V$ is the output vocabulary\footnote{\label{footnote:external_lm} Unlike our work, \cite{Variani20,zhang2021tiny} use phoneme labels instead of wordpieces, and an n-gram LM, as well as a lexicon, during first-pass decoding.}. Various other works adopting limited-context prediction networks with 3-4 history tokens are \cite{prabhavalkar2020less, zhang2020transformer} and \cite{zhang2021tiny}, the latter using a mere causal Conv1D layer over the history embeddings, and thus forgoing token-order information\footnotemark[\getrefnumber{footnote:external_lm}]. \cite{ghodsi2020rnn} goes further by just conditioning on a single previous label, making the prediction network function as a stateless embedding layer, and observes no degradation for low-resource languages. Nevertheless, relative WER regressions do occur for languages with large amounts of training data.

These works give clear indications that the size and complexity of the RNN-T encoder are more important for performance than those of the prediction network. This is supported by the findings in \cite{shrivastava2021echo}: When the prediction network is randomly initialized and frozen (not trained), WER never degrades compared to the fully-trained baseline. In contrast, similar freezing of the encoder layers hurts performance significantly.

In this work, we explore some useful changes in architecture for both the prediction and joint networks, that eliminate the performance gap to the full-context LSTM baseline \cite{shrivastava2021echo,prabhavalkar2020less}, while using a much smaller RNN-T decoder. First, we design our prediction network to be a weighted averaging of the history-token embeddings, where the weights are determined by a multi-headed attention mechanism that only attends to so-called \emph{position vectors}, without any cross-token attention. In addition, we tie the label embedding matrix to the output layer of the joint network, which is analogous to a common practice in LMs \cite{inan2016tying}. On a Voice Search task, we find that the proposed tied and reduced RNN-T decoder, at 2M parameters, has no loss in accuracy compared to the 23M parameter LSTM baseline decoder, if word-level Edit-based Minimum Bayes Risk (EMBR) training \cite{prabhavalkar2018minimum}\cite{weng2019minimum}\cite{guo2020efficient} is utilized. In addition, the proposed architecture is non-recurrent, fast and accelerator-friendly. The proposed decoder improves inference speed up to 3.7x on CPU.

%% file: rnnt.tex
\section{RNN-T and Embedding Decoders} \label{sec:rnnt}
\begin{figure}[t]
  \centering \includegraphics[scale=0.5]{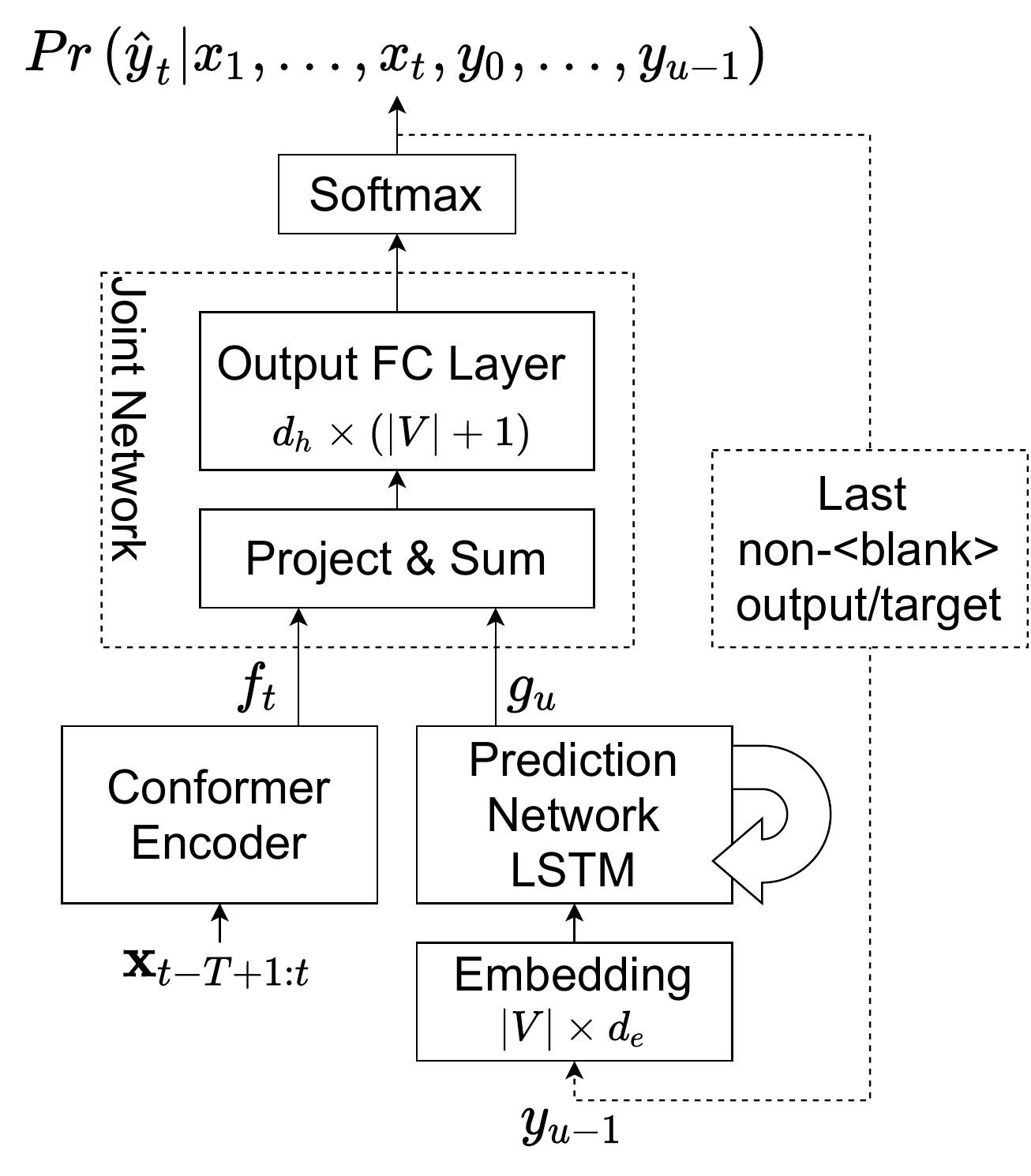}
  \caption{RNN-T baseline model with some weight dimensions.}
   \label{fig:rnnt}
   \vspace{-2mm}
\end{figure}

\newcommand{\blank}{{<}b{>}}

The RNN-T is a streaming E2E model \cite{graves2012sequence}, illustrated in Figure \ref{fig:rnnt}.  Its task is to produce a label sequence $\mathbf{y}$ (wordpieces in this work), given a stream of acoustic frames. At each time step $t$, the model receives a new acoustic frame $x_t$ and  outputs a probability distribution over $\hat{y}_t\in V \cup \{ \blank \}$, $V$ being the wordpiece vocabulary, and $\blank$ a blank symbol. The blank symbol is needed since the length of the label sequence is usually much shorter than the total number of acoustic frames. Hence, during training, the model attempts to learn to output sequences $\mathbf{\hat{y}}$ that would be identical to $\mathbf{y}$ if all $\blank$ outputs are removed.

RNN-T consists of 3 main parts: an encoder, a prediction network (PN) and a joint network. At each time step, the encoder (here a series of Conformer \cite{gulati2020conformer} layers)  produces an encoding $f_t$, while attending to a series of $T$ most recent frames received so far $\mathbf{x}_{t-T+1:t}$. Every time the model outputs a non-blank token, it is fed back into the PN to be embedded and passed through a series of recurrent LSTM layers. The PN thus produces a representation $g_u$, which can be conditioned on all previous non-blank outputs $y_0 \ldots y_{u-1}$, and is used for subsequent time steps until the next PN update. Inside the joint network, both $f_t$ and $g_u$ are projected to having the same dimensionality, and their sum goes through a fully-connected layer to produce the output logits.

\subsection{Limited-Context RNN-T}
In vanilla RNN-T models, the PN is a recurrent model conditioned on all previous non-blank predictions. On the other hand, a limited-history PN is only conditioned on the last $N$ such predictions. Works such that \cite{prabhavalkar2020less,zhang2020transformer} use LSTMs or Transformers with truncated history. Our study adopts limiting the history, but also focuses on ways to replace the PN architecture itself with smaller, non-recurrent and on-device friendly layers.

\subsection{Embedding Decoders}
\label{subsec:emb_decoders}

\cite{Variani20} suggests that, after training, the PN's LSTM conditioning history can be restricted to a size $N=2$ without loss in accuracy, and can thus be converted to a lookup table. A variation on this can be to train the lookup table directly. Since that work uses phonemes, size considerations would be different in our case: With thousands of wordpieces, the $V^2$ lookup table can be in the order $10^8$ parameters, i.e. too large for our purposes. One solution is to embed each token with a shared embedding matrix and concatenate the embeddings. We call decoders with such prediction networks Embedding Decoders, since their size and complexity are dominated by the embedding operation.

%% file: architecture.tex
\section{Reduced Embedding Decoders \label{sec:architecture}}
A downside of the Embedding Decoders from Section \ref{subsec:emb_decoders} is that, for longer histories $N$, concatenated embeddings require larger subsequent projection layers. Moreover, a mere embedding operation might not be expressive enough, esp. for small embedding dimensions. We design a more expressive layer that also cut down overall model size.

\subsection{Prediction Network Design}
\label{subsec:reduced}
\begin{figure}[t]
  \centering \includegraphics[scale=0.5]{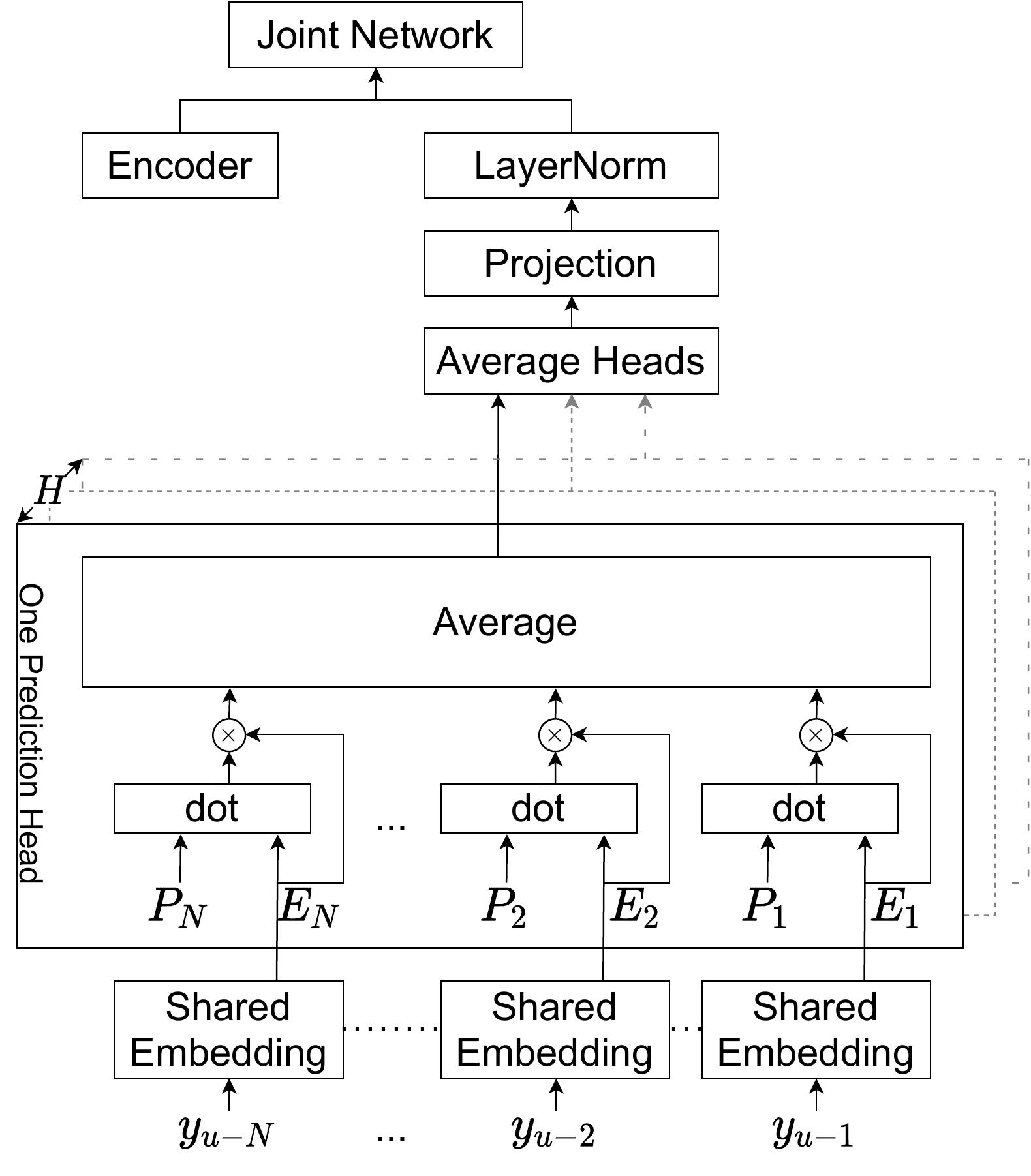}
  \caption{Proposed prediction network with $N$ previous tokens.}
   \label{fig:prediction network}
  \vspace{-2mm}
\end{figure}

Figure \ref{fig:prediction network} illustrates our prediction network. It uses a shared embedding matrix to embed each previous label $y_{u-i}$ into an embedding vector $E_i$. The aim of our design is to average rather than concatenate the embeddings, hence the name \emph{reduced}, and get a slimmer output. In order to still retain information about token order, we use so-called position vectors $P$: For each position in the history of our $N$ previous labels, we create a $P$ with the same size as the embedding. These are independent vectors, that can either be learned or set to some constant values. The output of our PN is the weighted average of all embeddings, where the averaging weight for each embedding is the dot product between itself and its position's $P$. This can be interpreted as attending over each embedding in proportion to its relevancy to the $P$ of its respective position. Thus, for

\newenvironment{nalign}{
    \vspace{-3mm}
    \begin{equation*}
    \begin{aligned}
}{
    \end{aligned}
    \end{equation*}
    \ignorespacesafterend
}

\begin{nalign}
\label{eq:one_head}
    \text{Embeddings } \mathbf{E} &\in \mathbb{R}^{N \times d_e}, \text{ Position Vectors } \mathbf{P} \in \mathbb{R}^{N \times d_e}, \\
    \text{Prediction}(\mathbf{E}, \mathbf{P}) &= \frac{1}{N } \sum_{n} \left[ \mathbf{E}_{n} \cdot \sum_e E_{n,e} \cdot P_{n,e}  \right] ,
\end{nalign}
where $n$ and $e$ respectively index the history positions and the embedding dimension. The interaction with $\mathbf{P}$ can potentially preserve position information, without needing recurrence or cross-token attention. Early experiments show that $\mathbf{P}$ don't need to be trainable, so we fix them to random values.

Inspired by \cite{vaswani2017attention}, we find that expanding this to a multi-headed design improves performance significantly. Each head has its own set of position vectors, but utilizes the same shared embedding matrix, costing just a small parameter increase overall. We average the outputs from all heads. For $H$ heads, indexed by $h$, and $\mathbf{P'} \in \mathbb{R}^{H \times N \times d_e}$, we get

\begin{nalign}
    \text{MultiHeadPred}(\mathbf{E}, \mathbf{P'}) &= \frac{1}{H \cdot N} \sum_{h,n} \left[ \mathbf{E}_{n} \cdot \sum_{e} E_{n,e} \cdot P'_{h,n,e} \right]
\end{nalign}

To further improve model expressiveness, we add a cheap projection layer, stabilized with LayerNorm \cite{ba2016layer}, followed by a Swish non-linearity \cite{ramachandran2017searching}. We apply those to the averaging output before passing it to the joint network, as shown in Figure \ref{fig:prediction network}. The extra projection layer does not alter the output dimension.

Overall, we get a PN with an output dimension equal to one embedding vector, achieving a slim architecture for the whole decoder. It remains small for any context size: For any increase in $N$, we just need some additional $P$ vectors. This has enabled us to experiment with longer histories $N>2$.

\section{RNN-T Decoder with Tied Embeddings}
\label{subsec:tying}
The works cited above have optimized the size of the prediction network without reducing the size of the joint network, which has a constant cost of 2-3M parameters. To further reduce the RNN-T decoder as a whole, we explore parameter sharing between the joint and the PN. As shown in Figure \ref{fig:rnnt}, the dimensions of the PN's embedding matrix are $|V| \times d_e$. Meanwhile, if the joint network's last hidden layer is of size $d_h$, the feed-forward projection weights from there to the output logits will have a shape of $d_h \times |V+1|$, with 1 extra output for the blank token. If we restrict our model to have $d_e=d_h$, these two matrices can share their weights for all non-blank tokens, after utilizing a simple transpose transformation. This is analogous to the weight tying practice in LMs \cite{inan2016tying}, and saves $d_h \cdot |V|$ parameters, while potentially regularizing the model.

%% file: experiments.tex
\section{Experimental Details \label{sec:experiments}}

\subsection{Data Sets}
\label{subsec:datasets}

The training set used for experiments is the same data from \cite{narayanan2019recognizing}. The multi-domain data covers domains of search, farfield, telephony and YouTube. All datasets are anonymized and hand-transcribed; the transcription for YouTube utterances is done in a semi-supervised fashion \cite{liao2013large}\cite{soltau2016neural}. To increase data diversity, we augment our dataset with multi-condition training (MTR) \cite{kim2017generation} and random 8kHz down-sampling \cite{li2012improving}.

The test set is also anonymized, hand-transcribed, and is representative of Google’s Voice Search traffic. It consists of around 12K utterances  with an average duration of 5.5 seconds. For long-form test sets, we use the YouTube and the TTS-Audiobook test sets from \cite{narayanan2019recognizing}. They have an average utterance length of 319 and 66 seconds, respectively.

\subsection{RNN-T Model Specifications}

All experiments use 128-dimensional log-Mel features, computed with a 32ms window and shifted every 10ms. Similar to~\cite{narayanan2019recognizing}, features for each frame are stacked with 3 frames to the left and then downsampled by three to a 30ms frame rate. Our model is similar to the one described in \cite{bo21system}. Thus, we use a 12-layer causal Conformer encoder \cite{gulati2020conformer} with 113M parameters.

We apply Specaugment \cite{park2019specaugment} in the manner described in \cite{park2020specaugment}. The model predicts 4,096 wordpiece units, including the end-of-sentence token. Training is done on 8x8 Cloud TPU using the Tensorflow Lingvo toolkit \cite{shen2019lingvo}. Finally, the 1st-pass output is rescored with a maximum-entropy LM \cite{biadsy2017effectively} in the second pass. When used, EMBR training is performed after normal training, for around one tenth the number of training steps.

\subsection{RNN-T Decoders in Comparison}

\renewcommand{\arraystretch}{1.15}

Table \ref{table:specs} gives some details about the different RNN-T decoders that we use in our experiments. The Size column shows the number of parameter for the prediction and joint networks put together. The \texttt{LSTM} baseline is the largest and only recurrent model. It has 2 LSTM layers with 2,048 hidden units and a 640-dimensional projection per layer. \texttt{Stateless1Emb} is the simplest Embedding Decoder, from \cite{ghodsi2020rnn}, where the PN is a mere embedding of one token. \texttt{Concat2Emb} is the Embedding Decoder from Section \ref{subsec:emb_decoders}. Those 3 baselines are compared to our proposed decoders \texttt{ReducedLarge} and \texttt{ReducedSmall}, which only differ in embedding dimension $d_e$ and length of conditioning history $N$. Both have number of heads $H=4$.

\begin{table} [h!]
    \centering
  \caption{Specifications for RNN-T decoders in our experiments.}
  \begin{tabular}{l|c|c|c} \toprule
    Decoder Name & \thead{Embedding \\ Dimension ($d_e$)} & \thead{History \\ Length ($N$)}  & Size \\ \Xhline{3\arrayrulewidth}
    \texttt{LSTM} & 128  & $\infty$ & 23M \\ \hline
    \texttt{Stateless1Emb} & 640  & 1 & 6.0M \\ \hline
    \texttt{Concat2Emb} & 640  & 2 & 6.4M  \\ \Xhline{3\arrayrulewidth}
    \texttt{ReducedLarge} & 1280  & 2 & 9.2M\\ \hline
    \texttt{ReducedSmall} & 320  & 5 & \textbf{1.9M} \\ \bottomrule
  \end{tabular}
  \label{table:specs}
\end{table}

%% file: results.tex
\section{Results \label{sec:results}}

\subsection{Result Overview}
\label{subsec:result_overview}

Table \ref{table:best_results} compares the pre- and post-EMBR WER for all decoders. We confirm that the \texttt{Concat2Emb}, inspired by \cite{Variani20}, has competitive performance, including when the $V^2$ lookup table is trained directly. \texttt{Stateless1Emb} is significantly worse, but gains a substantial improvement after EMBR training.
Our larger setup, \texttt{ReducedLarge} reaches the baseline WER of 6.1\%, even without EMBR. Our small setup, \texttt{ReducedSmall} with 1.9M parameters, benefits from an expansion of history size to $5$ and reaches the baseline \texttt{LSTM} WER after EMBR. Below, a series of ablation studies show how different parameter choices effect the performance of the reduced models.

\begin{table} [h!]
  \caption{WER for baselines and our \texttt{Reduced} models.}
  \centering
  \begin{tabular}{l|c|c|c} \toprule
  Decoder Name  & Size & \thead{Pre-EMBR\\WER} & \thead{Post-EMBR\\WER} \\ \Xhline{3\arrayrulewidth}
    \texttt{LSTM}  & 23M & \textbf{6.1\%} & \textbf{6.1\%} \\ \hline
    \texttt{Stateless1Emb}  & 6.0M & 6.6\% & 6.2\% \\ \hline
    \texttt{Concat2Emb}  &  6.4M & 6.2\% & 6.2\% \\ \Xhline{3\arrayrulewidth}
    \texttt{ReducedLarge} & 9.2M & \textbf{6.1\%} & \textbf{6.1\%} \\ \hline
    \texttt{ReducedSmall}  & \textbf{1.9M} & 6.4\% & \textbf{6.1\%} \\ \bottomrule
  \end{tabular}
  \label{table:best_results}

\end{table}

\subsection{Ablation Studies}
\subsubsection{Effect of Tying and Decoder Size}

\begin{figure}[h!]
  \centering \includegraphics[scale=0.5]{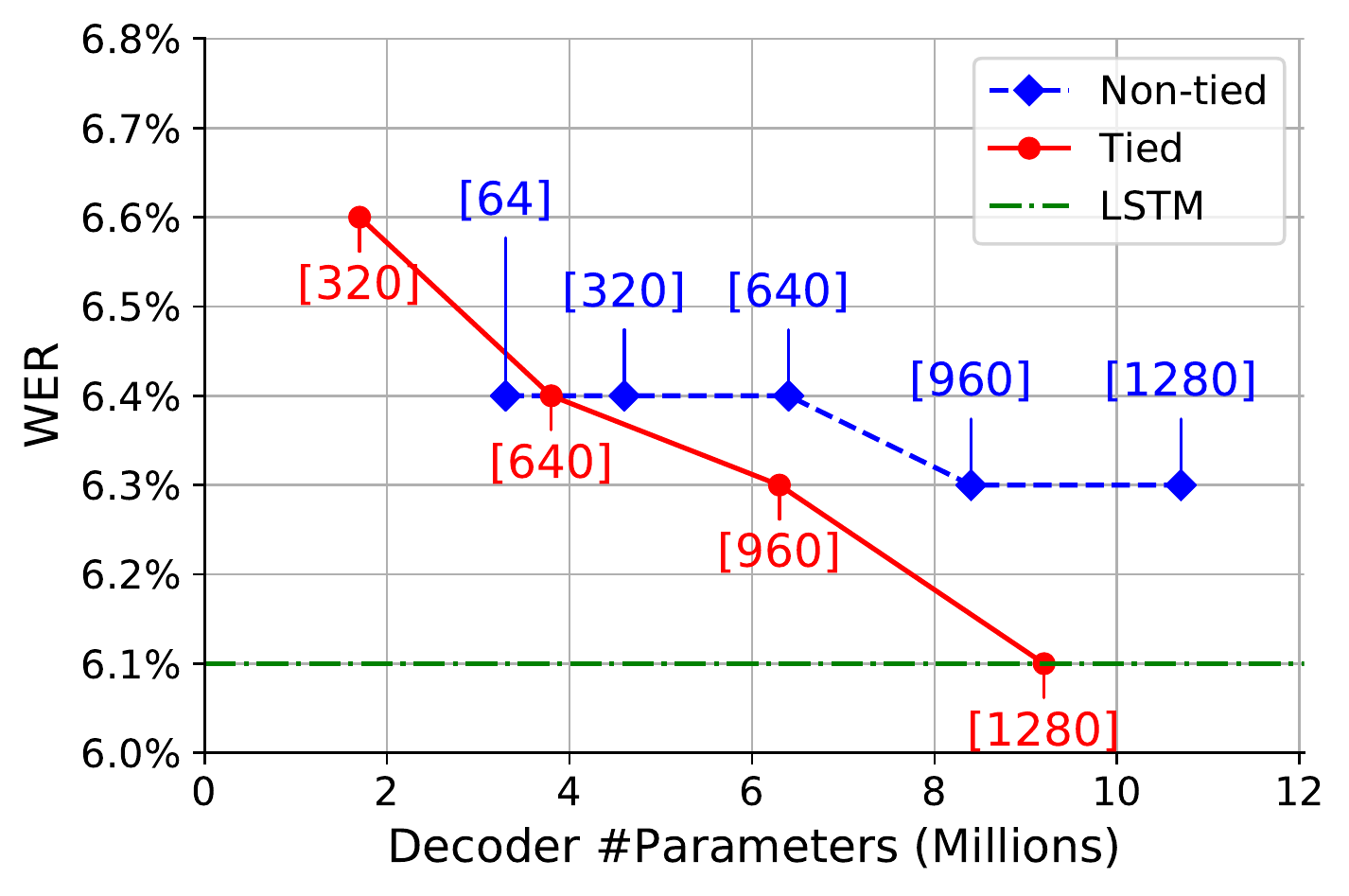}
  \caption{Pre-EMBR WER vs. size, for tied vs. non-tied reduced decoders. Embedding dimension $(d_e)$ are shown. All models here have 4 heads ($H$) and 2 history tokens ($N$).}
   \label{fig:wer_vs_size}
\end{figure}

Figure \ref{fig:wer_vs_size} shows how pre-EMBR WER varies for our reduced decoder with different sizes, with and without the tied embeddings from Section \ref{subsec:tying}. The sweep over model size is done by varying the embedding dimension $d_e$. Note that, in the non-tied case, the last hidden layer of the joint network is always set to be $d_h=640$. In the tied case, $d_h=d_e$, so the overall model size varies more with $d_e$. Moreover, however small $d_e$ becomes, the non-tied decoder will always have a size above 2.5M.

The results indicate that, starting from a decoder size of around 4M parameters, the weight-tying technique achieves better performance per model size. Moreover, for large-enough models, it reaches the same performance as the LSTM decoder, which the non-tied model never does. This might be a regularization effect. Below 4M parameters, the tied model rapidly loses performance. In the next subsection, we show how this can be mitigated by increasing the number of history tokens $N$.

\subsubsection{Varying The Number of History Tokens}

Table \ref{table:context_size} shows the pre-EMBR WER for \texttt{ReducedSmall} with different history lengths $N$, which hardly changes model size. Extending $N$ to 5 seems to offset the model's limited capacity.

\begin{table} [h!]
    \centering
    \caption{Pre-EMBR WER vs. number of history tokens ($N$) on \texttt{ReducedSmall} }
    
    \begin{tabular}{c|c|c|c|c}\toprule
    History length ($N$)   &3       &4       &\textbf{5}       &6 \\ \hline
    WER &6.6\%   &6.5\%   &\textbf{6.4\%}  &6.4\% \\ \bottomrule
    \end{tabular}
    \label{table:context_size}
    
\end{table}

\subsubsection{Varying Number of heads}

Table \ref{table:heads} shows Pre-EMBR WER when varying the number of heads for \texttt{ReducedLarge}. We observe that the extra heads seem to successfully augment our model with specialized position vectors. We choose $H=4$, since that works best with our other reduced models and is in agreement with the number of heads used in the encoder-decoder architecture in \cite{chiu2018state}.
\begin{table} [h!]
    \centering
    \caption{Pre-EMBR WER vs. \#heads ($H$) on \texttt{ReducedLarge}}
    \begin{tabular}{l|c|c|c|c|c|c}\toprule
        Heads   &   1       &   2       &   3       &   \textbf{4}       &   5       &   6 \\ \Xhline{2\arrayrulewidth}
        WER     &   6.2\%   &   6.2\%   &   6.1\%   &   \textbf{6.1\%}   &   6.1\%   &   6.3\% \\ \bottomrule
    \end{tabular}
    \label{table:heads}

\end{table}



\subsection{Long-Form Testsets}
Now that the parameters for \texttt{ReducedSmall} have been better understood, we take a closer look on how the various decoders perform on long-utterance test sets, post EMBR, as shown in Table \ref{table:long_utterances}. All Embedding Decoders perform significantly better than the \texttt{LSTM} baseline with its unbound context. Our proposed \texttt{ReducedSmall} decoder gives a further performance improvement over the baseline Embedding Decoders.

The LSTM has a long recurrent state dependency, and might be suffering from state saturation with the long utterances \cite{graves2012supervised}. While the \texttt{Statless1Emb} and \texttt{Concat2Emb} decoders do not have this recurrency problem, they have a limited history of 1-2 tokens and do not scale gracefully in size with a larger history like \texttt{ReducedSmall}. 

\begin{table} [h!]
  \centering
  \caption{Post-EMBR WER on long-utterance test sets}
  \begin{tabular}{l|c|c} \toprule
    Decoder Name &  Audiobook WER & YouTube WER \\ \Xhline{3\arrayrulewidth}
    \texttt{LSTM} & 5.0 \% & 10.5\% \\ \hline
    \texttt{Stateless1Emb} & 4.5\% & 10.3\% \\ \hline
    \texttt{Concat2Emb} & 4.4\% & 10.3\% \\ \Xhline{3\arrayrulewidth}
    \texttt{ReducedSmall} & \textbf{4.1\%} & \textbf{10.1\%} \\ \bottomrule
  \end{tabular}
  \label{table:long_utterances}

\end{table}

\subsection{Speedup measurement}
Table \ref{table:runtimes} compares the runtime statistics for the \texttt{LSTM} and \texttt{ReducedSmall} decoders on Pixel 4's Kyro 485 CPUs. We use 100 utterances (2.5 $\pm$ 1 s), running each 100 times.  Our smaller decoder achieves 2.7-3.7x inference speedup.

\begin{table} [h!]
  \centering
  \caption{Average Runtime Comparison for 10K runs}
  \begin{tabular}{l|c|c} \toprule
    CPU Core &  \texttt{LSTM} & \texttt{ReducedSmall} \\ \Xhline{3\arrayrulewidth}
    A55, 1.78 GHz & 19.2 $\pm$ 0.36 ms & \textbf{5.21 $\pm$ 0.15 ms} \\ \hline
    A76, 2.42 GHz & 2.23 $\pm$ 0.65 ms & \textbf{0.83 $\pm$ 0.43 ms} \\ \hline
    A76, 2.84 GHz & 2.03 $\pm$ 0.95 ms & \textbf{0.66 $\pm$ 0.03 ms} \\ \bottomrule
  \end{tabular}
  \label{table:runtimes}

\end{table}

%% file: conclusions.tex
\section{Conclusions \label{sec:conclusions}}
In this paper, we presented a design for a small, fast \mbox{RNN-T} decoder. Reducing the vanilla RNN-T decoder size by around 90\%, our 2M-parameter design is 3-4 times faster, and can completely fit in-memory on many on-device accelerator chips. We found that EMBR training is highly beneficial for small Embedding Decoders, helping our smallest model match or surpass baseline WER. Overall, our work presents further evidence that the size of the encoder is more important than that of the prediction network.
We also demonstrated that the common embedding-to-output weight sharing from language modeling is useful for RNN-T. It saved millions of parameters and achieved better WER than our best non-tied models. Future research can investigate whether the technique can also improve RNN-T with complex LSTM and/or Transformer decoders. We also recommend utilizing smaller decoders for complex beam search to further improve WER. Another direction enabled by our work is delivering multiple prediction networks on device, perhaps for model specialization or for multilingual models.